%% file: main.tex
\documentclass[10pt,twocolumn,letterpaper]{article}
\usepackage{cvpr}
\input{preamble}
\definecolor{cvprblue}{rgb}{0.21,0.49,0.74}
\usepackage[breaklinks,colorlinks,allcolors=cvprblue]{hyperref}
\hypersetup{
  pdftitle={TacPilot: Does Touch Need to Be Everywhere?},
  pdfauthor={Yunzhen Zhang, Yuxiao Zhu, Jie Yang, Yufeng Miao, Bing Luo}
}

\def\confName{CVPR}
\def\confYear{2027}
\title{TacPilot: Does Touch Need to Be Everywhere?}
\author{
  Yunzhen Zhang$^{1}$\thanks{Equal contribution.}\quad
  Yuxiao Zhu$^{1}$\footnotemark[1]\quad
  Jie Yang$^{2}$\quad
  Yufeng Miao$^{2}$\quad
  Bing Luo$^{1}$\\
  $^{1}$Duke Kunshan University\qquad
  $^{2}$Helix Robotics
}

\begin{document}
\maketitle
\input{sec/0_abstract}
\suppressfloats[t]
\input{sec/fig_teaser}
\input{sec/1_intro}
\input{sec/fig_method_framework}

\input{sec/2_related}
\input{sec/3_method}
\input{sec/4_experiments}
\input{sec/5_conclusion}
\clearpage
\section*{LLM Usage Statement}
We used OpenAI GPT for drafting, editing, and research
discussions. The authors are responsible for the
manuscript's claims, accuracy, and integrity.
{
  \small
  \bibliographystyle{ieeenat_fullname}
  \bibliography{main}
}
\end{document}

%% file: preamble.tex
\usepackage{colortbl}



%% file: sec/0_abstract.tex
\begin{abstract}
Tactile feedback helps visual manipulation policies respond
to local contact, but changes in tactile observations do not
always require corrective motion, and sustained corrections
do not always require new tactile features.
We introduce TacPilot, a tactile residual framework that
separates representation updates from action intervention
around a frozen visual policy.
TacPilot uses a perception gate to determine when tactile
features need refreshing and updates only affected regions
while reusing unchanged features.
A separate intervention gate determines whether bounded
Cartesian and gripper residuals modify the nominal visual action.
This separation allows tactile perception and action correction
to follow different spatial and temporal demands during manipulation.
In offline expert-trajectory replay, local feature reuse
reduces encoded regions by 56.4\% relative to periodic full
encoding, with essentially unchanged action predictions.
In an exploratory simulation evaluation spanning two cylinder
sizes, TacPilot succeeds in 26 of 30 trials (86.7\%).
We further validate TacPilot on OpenArm v2 equipped with
Paxini tactile sensors across four contact-rich manipulation tasks.
\end{abstract}

%% file: sec/fig_teaser.tex
\begin{figure}[t]
  \centering
  \includegraphics[width=\linewidth]{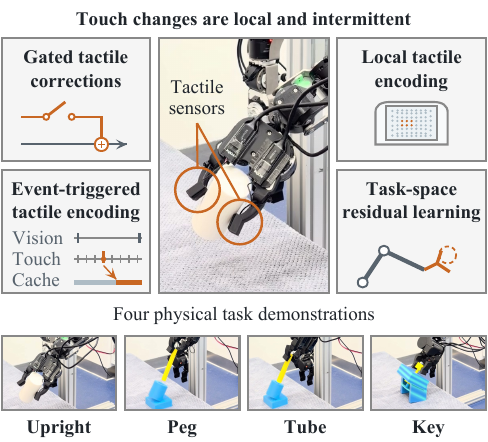}
  \caption{\textbf{TacPilot: spatiotemporal tactile sparsity.}
  Event-triggered local encoding and independently gated, bounded
  corrections complement supervised Cartesian residual learning.
  Photos show four physical tasks;
  diagrams are schematic.}
  \label{fig:teaser}
\end{figure}

%% file: sec/1_intro.tex
\section{Introduction}
\label{sec:intro}

Vision provides scene context and guides motion toward
a goal, while touch reveals local interactions that
can be difficult to observe under
occlusion~\cite{cheng2018active,zhan2025tactile}.
When uprighting a cylinder, for example, a robot may
need to adjust its grasp as the object shifts between
its fingertips. Tactile feedback provides information
about these contact changes that images alone may
not resolve. Recent work incorporates touch through
adaptive tactile injection~\cite{li2026atvla} and
tactile-conditioned residual policies~\cite{zhu2025equitac,
yu2026omnitactune,wang2026vitar}, enabling tactile feedback
to complement visual action generation.

Tactile interaction is local in space. In gripper-based
manipulation, contact with the object is concentrated
at the fingertip--object interface, where small changes
in grasp or alignment can determine whether a motion
succeeds~\cite{zhao2023act,zhan2025tactile}.
Within this interface, a shift in contact may change
the readings in only part of the sensing surface,
while other regions remain largely unchanged.
This locality suggests focusing corrective learning
on the end-effector interaction and reusing features
from unchanged regions.

The need for tactile feedback also varies over time.
Over a task, contact difficulties may require corrective
motion during some phases, while visual guidance is
sufficient during others. Within a shorter interval,
high-rate tactile sensing can capture contact changes
between successive visual observations, providing
feedback before the next visual update~\cite{li2026atvla}.
During stable interaction, however, many successive
tactile samples may carry little new information.
These variations motivate selective processing and
correction that retain access to rapid contact changes.

We study this spatial locality and temporal intermittency
as \emph{spatiotemporal tactile sparsity}
(Fig.~\ref{fig:teaser}).
We introduce TacPilot (Fig.~\ref{fig:method}), a tactile
residual framework that uses this structure to guide
feature updates and action intervention.
A frozen ACT policy~\cite{zhao2023act} supplies nominal
actions. A perception gate determines when tactile
features need refreshing, and local updates reuse
features from unchanged sensing regions.
An intervention gate controls whether bounded Cartesian
and gripper residuals modify the nominal action.
We express and supervise corrections in Cartesian action
space to focus learning on motion at the
gripper--object interface. The residual and intervention
gates are trained by behavior cloning while the visual
policy remains fixed.

We evaluate TacPilot on simulated cylinder uprighting
with two object sizes and variations in starting position
and grasp orientation. TacPilot succeeds in 26 of 30
development trials (86.7\%).
Offline ablations examine tactile refresh, residual
intervention, and input dependence. Selective updates
reduce the number of encoded regions by 56.4\% with
essentially unchanged action predictions.
We further validate TacPilot on the OpenArm v2 platform
across four contact-rich manipulation tasks.

Our contributions are threefold:
\begin{enumerate}
    \item \textbf{Spatiotemporal tactile sparsity.}
    We identify spatial locality and temporal intermittency
    as opportunities for selective tactile processing
    and action correction.

    \item \textbf{TacPilot.}
    We design TacPilot to exploit this sparsity through
    event-triggered local feature updates and gated
    task-space residual corrections.

    \item \textbf{Evaluation and real-robot validation.}
    We evaluate cylinder uprighting with two object sizes,
    examine the components through offline ablations,
    and validate TacPilot on OpenArm v2 across four
    contact-rich tasks.
\end{enumerate}

%% file: sec/fig_method_framework.tex
\begin{figure*}[t]
  \centering
  \includegraphics[trim=9.9bp 0bp 9.9bp 0bp,clip,width=\textwidth]{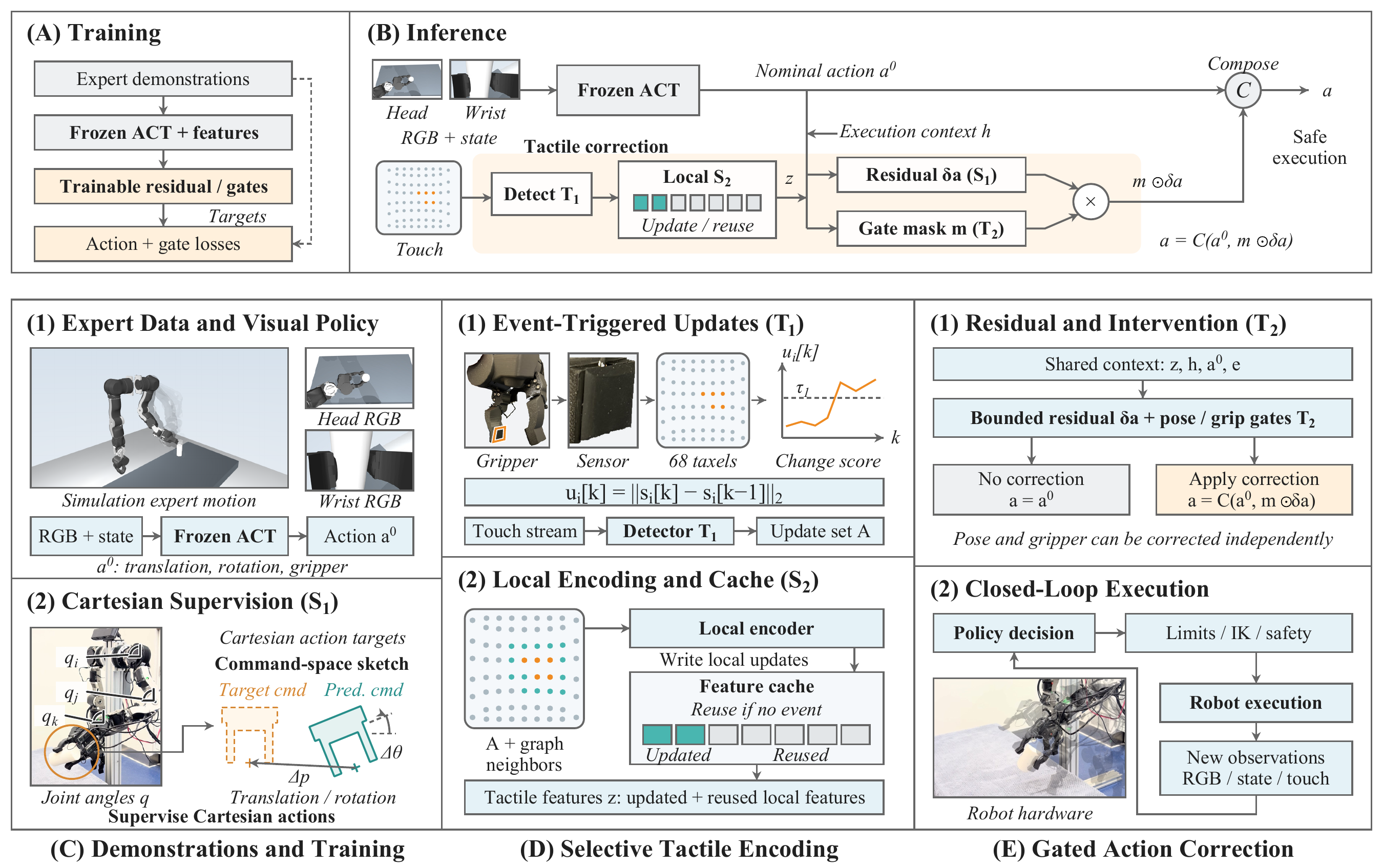}
  \caption{\textbf{TacPilot framework.}
  \textbf{(A) Training.}
  Behavior cloning trains the residual/gating head while ACT and inherited
  features remain frozen; dashed arrows denote supervision.
  \textbf{(B) Inference.}
  Bounded tactile residuals selectively correct nominal ACT actions $a^0$,
  using tactile features $z$, execution context $h$, and regional
  event summaries $e$ from the raw tactile monitor.
  \textbf{(C) Demonstrations and training.}
  Simulation demonstrations supply paired RGB/state inputs and Cartesian
  action targets ($S_1$); the target/predicted grippers depict command-space
  translation and rotation schematically.
  \textbf{(D) Selective tactile encoding.}
  $T_1$ triggers local feature updates ($S_2$), reusing valid cached tokens;
  the point layouts show 68 sensor-local taxel positions.
  \textbf{(E) Gated action correction.}
  Independent pose/gripper gates ($T_2$) control intervention, followed by
  shared action limits, IK, safety checks, and observation feedback.
  The mask $m$ acts component-wise; time subscripts are omitted.
  Faded poses come from simulation recordings or real video;
  highlights, offsets, and gate paths are schematic.}
  \label{fig:method}
\end{figure*}

%% file: sec/2_related.tex
\section{Related Work}
\label{sec:related}

\subsection{Visual--Tactile Manipulation}

ACT and Diffusion Policy learn manipulation behaviors
from visual demonstrations~\cite{zhao2023act,chi2023diffusion};
OpenVLA and $\pi_{0.5}$ extend visual policy learning
through large-scale pretraining~\cite{kim2024openvla,black2025pi05}.
Touch complements these policies with local contact information.
Prior work combines vision and touch for precision
grasping~\cite{matak2023planning} and tactile-reactive
control~\cite{xu2024letac}. See, Hear, and Feel uses
attention to fuse visual, auditory, and tactile histories
for manipulation~\cite{li2023seehearfeel}.
3D-ViTac places visual and tactile observations in a
shared 3D representation for diffusion-policy
learning~\cite{huang2025vitac}. NCF-v2 estimates
object--environment contacts from fingertip sensing and
uses these estimates to learn insertion
policies~\cite{higuera2023extrinsic}.

Temporal designs address the different rates of vision
and touch. Reactive Diffusion Policy combines a slow
latent action-chunk predictor with a fast decoder conditioned
on tactile or force feedback~\cite{xue2025reactive}.
AT-VLA learns when to inject tactile features and uses
slow visual-language and fast tactile streams for
reactive control~\cite{li2026atvla}. ViTaL separates
scene-level reaching from local visual--tactile
interaction~\cite{zhao2025vital}.
These methods establish spatial and temporal structure
as useful design choices. TacPilot studies selective
feature refresh and residual intervention within execution.

\subsection{Tactile Residual Policies}

Residual policy learning augments an existing controller
with learned corrections. Johannink et al.\ combine
feedback control with residual reinforcement learning
for contact-rich assembly~\cite{johannink2019residual}.
ResiP adds a closed-loop RL residual to a frozen
action-chunking BC policy for precise
assembly~\cite{ankile2025refinement}.
Tactile residual methods further condition corrections
on local contact observations. EquiTac uses tactile
normal maps and rotational equivariance to predict
orientation corrections~\cite{zhu2025equitac}.
OmniTacTune adapts frozen visual policies through
warm-start training followed by online tactile residual
RL~\cite{yu2026omnitactune}.
ViTaR uses outcome-based preferences to select bounded
residuals and tactile feedback to scale their
magnitude~\cite{wang2026vitar}.
TacPilot trains its residual and intervention gates
with behavior cloning and couples action selection
with event-triggered local feature reuse.

\subsection{Tactile Encoding and Selective Updates}

Tactile pretraining reduces dependence on task-specific
labels. T-DEX learns tactile representations from
robotic play for demonstration-based dexterous
manipulation~\cite{guzey2023dexterity}.
UniT studies data-efficient tactile
encoding~\cite{xu2025unit}; T3, AnyTouch, and Sparsh
learn representations across sensors or
tasks~\cite{zhao2025t3,feng2025anytouch,higuera2025sparsh}.
Masked Multimodal Learning jointly learns visual--tactile
representations and manipulation policies through
masked reconstruction and RL~\cite{sferrazza2024senses}.
Its masking objective operates during representation
learning.

Selective computation also exploits changes in a sensory
stream. NeuTouch and visual--tactile spiking networks
use event-based sensors for object classification and
slip detection~\cite{taunyazov2020event}.
In video inference, DeltaCNN propagates sparse frame
differences and reuses accumulated activations across
frames~\cite{parger2022deltacnn}.
TacPilot applies change-driven reuse to regional tactile
features from continuously sampled sensors, alongside
a gate controlling action intervention.

%% file: sec/3_method.tex
\section{Method}
\label{sec:method}

\subsection{Overview and Action Formulation}
\label{sec:method_overview}
\label{sec:cartesian_imitation}

TacPilot separates two decisions: when and where to update
tactile features, and whether to use them to correct an action.
A frozen ACT policy~\cite{zhao2023act} maps RGB $I_t$ and
robot state to a nominal action $a_t^0$ (Fig.~\ref{fig:method}B).
The tactile branch updates regional features ($T_1$, $S_2$)
and proposes bounded task-space corrections ($S_1$).
Pose and gripper gates ($T_2$) select the applied components.

\paragraph{Action coordinates.}
Each action $a_t\in[-1,1]^7$ contains three translation
increments, three rotation increments, and an absolute
gripper command. Cartesian correction and supervision
(Fig.~\ref{fig:method}C) directly describe motion at the
gripper--object interface; joint-space corrections produce
configuration-dependent displacements.

\paragraph{Available observations.}
At action decision $t$, the correction head receives tactile
features $z_t$, execution context $h_t$, the nominal action
$a_t^0$, and an event summary $e_t$.
Context $h_t$ combines robot state/history with object geometry
and a local goal from frozen RGB modules. The tactile monitor
summarizes the latest 50\,ms of raw touch as $e_t$, preserving
transients even when encoded features are reused.
The two tactile inputs serve complementary roles:
$z_t$ provides the encoded contact representation, while
$e_t$ exposes recent changes without waiting for a feature
refresh. Both use only observations available at decision time.

\subsection{Selective Tactile Encoding}
\input{sec/fig_main_results}
\label{sec:tactile_features}

\paragraph{When to refresh: $T_1$.}
A lightweight monitor detects changes directly from raw
touch (Fig.~\ref{fig:method}D). For taxel forces
$s_i[k]\in\mathbb R^3$ at sample $k$, region $\mathcal R_r$
is triggered by adjacent change or drift from its last
refresh reference $s_i^{\mathrm{ref}}$:
\begin{equation}
 \begin{aligned}
 c_r[k]&=\sum_{i\in\mathcal R_r}
                    \|s_i[k]-s_i[k-1]\|_2,\\
 d_r[k]&=\sum_{i\in\mathcal R_r}
                    \|s_i[k]-s_i^{\mathrm{ref}}\|_2,\\
 \mathcal A[k]&=\{r:\max(c_r[k],d_r[k])\geq\tau_1\}.
 \end{aligned}
 \label{eq:perception_gate}
\end{equation}
The drift term captures slowly accumulating changes;
the threshold $\tau_1$ is calibrated on training data.

\paragraph{Where to refresh: $S_2$.}
The encoder partitions two 68-taxel fingertips into
32 regions. Two spatial graph layers
precede temporal encoding~\cite{ning2026touchwgnn,bai2018tcn}.
Graph connections carry regional changes to neighboring
output tokens. We therefore expand the triggered set to
affected outputs, computing each with its two-hop dependencies.
Other tokens stay cached; the aggregation network combines
the available regional tokens into $z_t$.
Only completed updates enter the cache, and age limits
prevent indefinite reuse.

Closed-loop evaluation synchronously refreshes all regions
during contact.
Offline ablations instead skip encoding only for unchanged
input and graph-dependency windows (exact-window reuse).

\subsection{Gated Task-Space Correction}
\label{sec:intervention}

\paragraph{Bounded residual proposal.}
A trainable event-conditioned head adjusts the residual
and gate outputs of a frozen pretrained predictor using
its representation and $e_t$. Denoting the combined
proposal function by $f_\theta$,
\begin{equation}
 \begin{aligned}
 \widehat a_t&=a_t^0+B\odot\tanh f_\theta(z_t,h_t,a_t^0,e_t),\\
 \widetilde a_t&=\operatorname{clip}(\widehat a_t,-1,1),\\
 \delta a_t&=\widetilde a_t-a_t^0,
 \end{aligned}
 \label{eq:residual}
\end{equation}
where $B=(0.25\mathbf1_6,2)$ bounds the normalized
proposal. Clipping enforces action bounds; subtracting the
nominal action gives the feasible correction $\delta a_t$.

\paragraph{When to intervene: $T_2$.}
The combined predictor also produces pose and gripper logits
$r_t=R_\phi(z_t,h_t,a_t^0,e_t)\in\mathbb R^2$.
At inference, their probabilities are thresholded:
\begin{equation}
 \begin{aligned}
 b_t^p&=\mathbf1[\sigma(r_t^p)\geq0.5],\\
 b_t^g&=\mathbf1[\sigma(r_t^g)\geq0.5],\\
 m_t&=(b_t^p\mathbf1_6,\ b_t^g).
 \end{aligned}
 \label{eq:intervention_gate}
\end{equation}
Pose gating controls all six motion components; gripper
gating controls opening/closing. Before execution constraints,
composition $\mathcal C$
(Fig.~\ref{fig:method}B,E) gives
\begin{equation}
 a_t^{\mathrm{learn}}=a_t^0+m_t\odot\delta a_t.
 \label{eq:gated_composition}
\end{equation}
A closed gate preserves the corresponding ACT components.
Feature refresh and intervention are independent:
new features can leave the nominal action unchanged,
and cached features can support continued correction.

\subsection{Training and Closed-Loop Execution}
\label{sec:residual_training}

\paragraph{Joint behavior cloning.}
We train the head on expert Cartesian commands $a_t^\star$
(Fig.~\ref{fig:method}A,C), keeping ACT,
the tactile encoder, visual goal modules, and inherited
representation and predictor frozen.
Gate labels $y_t$ threshold expert--nominal command
discrepancies normalized by $B$ at 0.02: any of the six
pose components activates its label; gripper is labeled separately.
Training uses $p_t=\sigma(r_t)$ instead of binary gates in
Eq.~\ref{eq:gated_composition}, forming a differentiable
soft action $\bar a_t$.
The objective is
\begin{equation}
 \begin{aligned}
 \mathcal L_{\mathrm{BC}}={}&
 2\ell_B(\widehat a_t,a_t^\star)
 +2\ell_B(\bar a_t,a_t^\star)\\
 &+\operatorname{BCE}(p_t,y_t^\epsilon)
 +\tfrac12(\mathcal L_{\mathrm{temp}}+\mathcal L_{\mathrm{next}}),
 \end{aligned}
 \label{eq:bc_loss}
\end{equation}
where $\ell_B$ is Smooth L1 command error normalized by
the correction bounds, and $y_t^\epsilon$ denotes smoothed
gate labels. The first two terms fit the unclipped proposal
and soft composed action; binary cross-entropy supervises intervention.
Auxiliary losses match expert command changes and next
commands on adjacent recorded pairs. The current prediction
replaces command history for the next prediction;
observations remain fixed.

\paragraph{Execution and feedback.}
Hard-gated commands pass through action limits, IK, and
safety checks (Fig.~\ref{fig:method}E). After a confirmed grasp,
a separate release-memory rule holds the gripper open when
both fingers unload, the gripper is open, and an opening
command persists for three decisions.
Applied actions and observations feed the next decision.

%% file: sec/fig_main_results.tex
\begin{table*}[t]
  \centering
  \small
  \setlength{\tabcolsep}{0pt}
  \begin{tabular}{@{}p{0.25\textwidth}*{3}{>{\centering\arraybackslash}p{0.17\textwidth}}*{2}{>{\centering\arraybackslash}p{0.12\textwidth}}@{}}
    \toprule
    Method & $4\times8$\,cm $\uparrow$ & $5\times10$\,cm $\uparrow$ & Overall $\uparrow$ & Safety stops $\downarrow$ & Timeouts $\downarrow$\\
    \midrule
    Vision-only ACT & 7/16 (43.8\%) & 8/14 (57.1\%) & 15/30 (50.0\%) & 9/30 & 6/30\\
    Dense tactile residual & \textbf{13/16 (81.3\%)} & 8/14 (57.1\%) & 21/30 (70.0\%) & 6/30 & 3/30\\
    \rowcolor[RGB]{225,239,250}
    TacPilot (ours) & \textbf{13/16 (81.3\%)} & \textbf{13/14 (92.9\%)} & \textbf{26/30 (86.7\%)} & \textbf{2/30} & \textbf{2/30}\\
    \bottomrule
  \end{tabular}
  \caption{\textbf{Success rates and failure outcomes for cylinder uprighting.}
  We compare ACT, DTR, and TacPilot on two cylinder sizes
  (diameter $\times$ length). Success requires the cylinder to remain standing
  after release and withdrawal. Success columns show successes/trials
  (percentage); safety stops and timeouts show counts/trials.
  Blue highlights TacPilot, and bold marks the best values, including ties.}
  \label{tab:main_results}
\end{table*}

%% file: sec/4_experiments.tex
\section{Experiments}
\label{sec:experiments}

\input{sec/fig_matched_comparison}
\input{sec/tab_ablations}
\input{sec/fig_core_analysis}
\input{sec/fig_physical_validation}

We evaluate complete-task performance in simulation,
analyze selective perception and intervention on recorded
histories, and qualitatively validate TacPilot across four
physical tasks on OpenArm v2.

\subsection{Tasks and Evaluation Protocol}
\label{sec:experiment_setup}

\paragraph{Simulation setup.}
We use MuJoCo to simulate OpenArm v2 with modeled Paxini
tactile pads on the gripper. Finger--object contact forces
are mapped to a sensor-local layout of 68 taxels per
fingertip, providing three-axis simulated tactile readings.
Head and left-wrist cameras provide rendered RGB observations.

\paragraph{Cylinder uprighting.}
The robot grasps a lying cylinder, brings it upright, releases it
on the table, and withdraws while the object remains standing.
This sequence requires both establishing contact and ending it
without disturbing the object.
We evaluate $4\times8$\,cm and $5\times10$\,cm cylinders
(diameter $\times$ length), with 16 and 14 starts.
Position offsets span $\pm2$\,cm and $\pm5$\,cm along the
table-plane axes; grasp pitch ranges from $-30^\circ$ to
$-10^\circ$. Material, friction, lighting, and control delay
remain fixed. Diameter and length vary together, and object
tilt is not randomized.

The 30 starts combine 19 previously evaluated starts with
11 additional seeded starts, using preassigned size quotas.
Because earlier starts were reused and the size range was
restricted after earlier experiments, this is an exploratory
development evaluation. Success requires stable standing after
release and withdrawal; timeouts and safety stops count as
failures. The tactile methods share initial case parameters
and 20\,Hz action decisions.
All methods use a 2\,mm penetration-stop threshold.

\paragraph{Compared systems.}
\textbf{Vision-only ACT} (Action Chunking with
Transformers)~\cite{zhao2023act} is a behavior-cloning policy
that predicts future Cartesian and gripper command sequences
from head/left-wrist RGB and robot state. These commands
are executed without tactile input or residual correction.
\textbf{Dense tactile residual (DTR)} is our reimplementation
of OmniTacTune~\cite{yu2026omnitactune}. It augments a frozen
ACT policy with dense tactile encoding and an always-on
correction, trained with online soft actor-critic (SAC).
\textbf{TacPilot} uses selective feature updates and independent
pose/gripper gates; its joint event head is trained by BC with
ACT and inherited feature networks frozen.
Only TacPilot uses the release-memory rule in
Sec.~\ref{sec:residual_training}; ACT and DTR do not.

\subsection{Comparison with Baselines}
\label{sec:main_comparison}

We evaluate TacPilot against the vision-only ACT and DTR
baselines using task success rates across the two object sizes
(Table~\ref{tab:main_results}).
TacPilot completes 26/30 trials (86.7\%), compared with
21/30 (70.0\%) for the dense tactile reference and
15/30 (50.0\%) for vision-only ACT.
The difference between the tactile systems is concentrated
at $5\times10$\,cm: TacPilot succeeds in 13/14 trials,
versus 8/14 for each reference.
The overall difference from the dense reference is 16.7
percentage points. The failure breakdown also favors TacPilot:
it records two safety stops and two timeouts, compared with
six safety stops and three timeouts for DTR, and nine safety
stops and six timeouts for ACT.
Both training and release memory differ between systems.

Figure~\ref{fig:matched_comparison} follows a selected case
from the same initial condition for all three policies.
Each policy grasps and transports the cylinder, but their
behavior diverges during placement and release.
Vision-only ACT still holds the cylinder in the final
frame (A), leaving release and withdrawal unfinished.
DTR releases before stable placement, and the object topples
(B). Both executions time out despite their contrasting
behavior at the end of contact.
TacPilot brings the cylinder into placement contact,
then releases and withdraws, leaving it standing (C).
The interaction and final frames show support transferring
from the gripper to the table. Successful completion requires
coordinating placement, gripper opening, and withdrawal.

\subsection{Ablation Study}
\label{sec:ablations}

\paragraph{Ablation setup.}
We now examine individual components by changing the feature
update rule, intervention gates, or tactile inputs while
keeping model weights and observations fixed
(Table~\ref{tab:ablations}). All conditions replay 12 expert
trajectories from seven internal model-selection starts,
totaling 14,865 frames, without retraining or new rollouts.
These offline comparisons measure tactile computation and
agreement with expert commands.
Normalized MSE balances starts within each phase and averages
the four phases equally ($\times10^3$ throughout).
Translation/rotation RMSE and gripper MAE measure command
errors, with gripper values in native $[-1,1]$ units.
Touch is sampled at 1,000\,Hz and actions at 20\,Hz;
gates use a 0.5 threshold without release memory.

\paragraph{Reducing tactile encoding while preserving predictions.}
Table~\ref{tab:ablations}(A) separates when features are
refreshed from how many regions are recomputed.
Periodic offers fixed-rate updates; Event uses the perception
gate $T_1$. Full encodes every region, while Local ($S_2$)
reuses features whose complete input and graph-dependency
windows are unchanged.
Starting from Periodic/Full, enabling event timing alone
(Event/Full) reduces encoded regions by 14.1\% and encoding
time by 14.5\%. Event/Local reduces regions by 56.4\% and
encoding time from 30.60 to 19.17\,s (37.3\%).
Action MSE stays near 0.948 across all four conditions,
and maximum action differences remain below $8\times10^{-7}$.
In the local conditions, unchanged windows can therefore
be reused while preserving the model's predictions.

Local reuse accounts for most of this reduction:
Periodic/Local and Event/Local both encode 206,975 regions
in 9,290 calls, despite different refresh-request counts.
Event/Local has the lowest encoding time, whereas
Periodic/Local has the lowest processing time (68.00\,s).
Times are synchronized A100 wall-clock measurements including
transfers; they exclude the shared event summary, actor,
simulation, file I/O, and model loading.

\paragraph{Pose and gripper corrections have different tradeoffs.}
Having examined feature computation, we keep the correction
proposals fixed and vary whether each component is applied
(Table~\ref{tab:ablations}(B)). Each row specifies the
pose/gripper setting: G uses the learned gate, and A applies
the correction continuously.
At either pose setting, switching the gripper from always-on
to gated reduces gripper MAE from 0.04193 to 0.00841
(79.9\%).
With the gripper gated, always-on pose (A/G) gives the
lowest overall MSE, 9.1\% below G/G, and lower translation/rotation errors.

The phase errors clarify this difference. Relative to G/G,
A/G improves alignment MSE from 2.898689 to 2.485711,
but raises approach error from 0.016246 to 0.060888.
Always-on gripper slightly improves release error while
worsening approach and alignment. Pose gating thus helps
approach at the cost of alignment accuracy.

\paragraph{Rich tactile features and short-term events contribute differently.}
We finally examine which tactile inputs support the command
predictions (Table~\ref{tab:ablations}(C)).
No rich removes the encoded tactile features $z_t$;
No events removes the short-term event summary $e_t$;
Neither applies both masks. Basic force/contact and
feature-validity context remain available in every condition.
Removing rich features raises overall MSE to $7.70\times$
the full-input reference, with the largest phase error
increase during release/withdrawal (0.575885 to 24.530964).
Rich features thus strongly affect predicted release commands.

Removing events has a smaller, phase-dependent effect:
overall MSE rises by 3.07\% and grasp-correction error by
18.2\%, while gripper MAE improves from 0.00841 to 0.00680.
The event summary helps some predictions but introduces
tradeoffs in others.
Only 31 frames cover grasp correction, and frames are
temporally dependent, limiting the evidence for that phase.

\subsection{Perception and Intervention Analysis}
\label{sec:perception_intervention_analysis}

Figure~\ref{fig:core_analysis} aligns an expert trajectory
and model predictions. Keyframes (A) identify approach, grasp,
rotation, placement, and release. In (B), rows represent the
32 fingertip regions, grouped by finger.
Each 50\,ms bin shows peak adjacent-sample force change:
darker colors indicate rapid changes rather than sustained loads.
Panel (C) shows the corresponding processing:
orange denotes recomputation, teal exact feature reuse,
and white no scheduled update. Recomputation also reflects
cumulative drift and graph dependencies, so dark regions in
(B) need not map one to one to orange regions in (C).

Panels (C) and (D) separate perception from intervention.
Many regions are recomputed
during contact after grasping, while the pose probability
remains below the 0.5 threshold for much of this interval.
Pose intervention becomes active toward placement, while
the gripper gate opens around grasping and again near release.
The threshold enables each correction independently.
Regional updates can thus continue with pose intervention disabled.

Panel (E) links pose intervention to translation/rotation
proposals and release to a negative gripper correction toward
opening. Separate normalization emphasizes timing; equal
curve heights do not imply equal physical magnitudes.
These proposals are unexecuted: the keyframes show expert
motion, not the outcome of applying the corrections.

\subsection{Real-World Experiments}
\label{sec:physical_validation}

\paragraph{Platform and tasks.}
Beyond simulated uprighting, TacPilot performs four physical tasks on
OpenArm v2 with Paxini tactile sensing
(Fig.~\ref{fig:physical}). RGB observations come from an Orbbec
Astra Pro head camera and a left-wrist UVC camera.
Task execution uses TacPilot without human intervention.
Each sequence combines external and head/wrist images
with fingertip tactile readouts.

\paragraph{Cylinder uprighting.}
The robot picks up the cylinder, turns it upright, and
places it on the table before opening and moving the
gripper away (A). The task requires a secure grasp during
lifting and rotation, followed by a release that leaves
the cylinder standing. This is the same complete-task
requirement used in simulation.

\paragraph{Peg and tube insertion.}
The robot brings the solid peg (B) or hollow tube (C)
to its fixture, aligns the part, and advances it into
place before withdrawing. These tasks require the grasp
to be maintained as the held part makes contact with
the fixture. The two sequences show this transition
for different object geometries.

\paragraph{Key manipulation.}
The robot engages the key, turns it, and then pulls it
out (D). The gripper must keep hold of the key as the
motion changes from rotation to extraction.

Together, these tasks span different contact transitions:
uprighting ends with release, while insertion and key
manipulation require continued grasping as the held object
contacts and moves relative to a fixture.

%% file: sec/fig_matched_comparison.tex
\begin{figure*}[t]
  \centering
  \includegraphics[width=\textwidth]{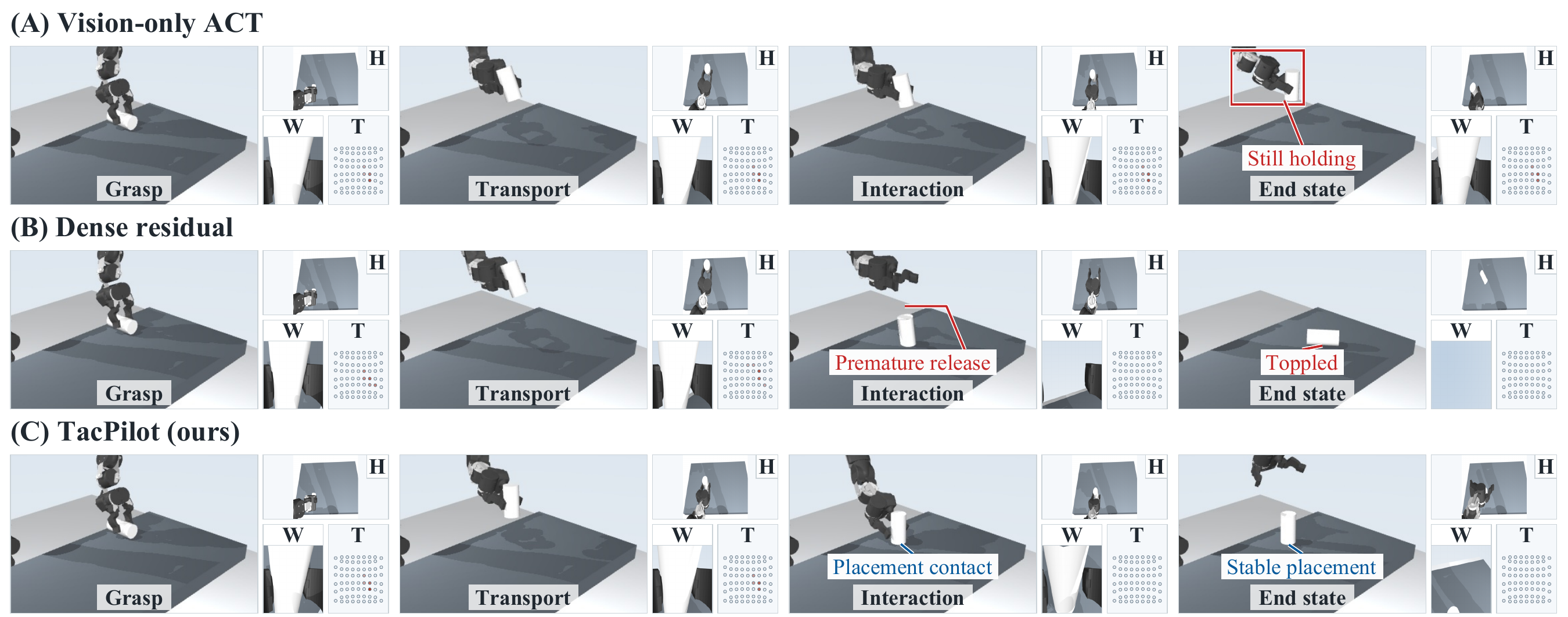}
  \caption{\textbf{Example upright executions from the same initial condition.}
  Rows show (A) Vision-only ACT, (B) Dense residual, and (C) TacPilot (ours).
  Each overview accompanies head RGB (H), wrist RGB (W), and an inner-pad
  68-taxel map (T). Colors encode simulated force-magnitude RMS over the
  trailing 100\,ms on a shared 0--9\,N square-root scale, not hardware
  measurements. Images are rendered by replaying the executed actions;
  they are not the RGB observations originally supplied to the policies.
  Vision-only T is passive and was not supplied to the policy.
  Stage times differ across methods. ACT and the dense residual time out;
  TacPilot succeeds.}
  \label{fig:matched_comparison}
\end{figure*}

%% file: sec/tab_ablations.tex
\begin{table*}[t]
  \centering
  \small
  \setlength{\tabcolsep}{0pt}
  \begin{tabular}{@{}p{0.14\textwidth}p{0.12\textwidth}>{\raggedleft\arraybackslash}p{0.11\textwidth}>{\raggedleft\arraybackslash}p{0.10\textwidth}>{\raggedleft\arraybackslash}p{0.12\textwidth}>{\raggedleft\arraybackslash}p{0.14\textwidth}>{\raggedleft\arraybackslash}p{0.17\textwidth}>{\raggedleft\arraybackslash}p{0.10\textwidth}@{}}
    \toprule
    \multicolumn{8}{l}{\textbf{(A) Refresh factorial: event timing ($T_1$) and local reuse ($S_2$)}}\\
    \mbox{Timing ($T_1$)} & \mbox{Reuse ($S_2$)} & Requests & \mbox{Calls $\downarrow$} & \mbox{Regions $\downarrow$} & \mbox{Encoder (s) $\downarrow$} & \mbox{Enc.\ (ms/call) $\downarrow$} & \mbox{Proc.\ (s) $\downarrow$}\\
    \midrule
    Periodic & Full & 14,841 & 14,841 & 474,912 & 30.60 & 2.06 & 79.12\\
    Event & Full & 12,753 & 12,753 & 408,096 & 26.16 & \textbf{2.05} & 76.86\\
    Periodic & Local & 14,841 & \textbf{9,290} & \textbf{206,975} & 19.82 & 2.13 & \textbf{68.00}\\
    \rowcolor[RGB]{225,239,250}
    Event & Local & 12,753 & \textbf{9,290} & \textbf{206,975} & \textbf{19.17} & 2.06 & 69.73\\
    \bottomrule
  \end{tabular}
  \par\smallskip
  \begin{tabular}{@{}p{0.07\textwidth}>{\raggedleft\arraybackslash}p{0.13\textwidth}*{2}{>{\raggedleft\arraybackslash}p{0.11\textwidth}}>{\raggedleft\arraybackslash}p{0.10\textwidth}*{4}{>{\raggedleft\arraybackslash}p{0.12\textwidth}}@{}}
    \toprule
    \multicolumn{9}{l}{\textbf{(B) Pose and gripper intervention ($T_2$): overall and phase-resolved errors}}\\
    & \multicolumn{4}{c}{Overall errors} & \multicolumn{4}{c}{Phase MSE ($\times10^3$)}\\
    \cmidrule(lr){2-5}\cmidrule(lr){6-9}
    P/G & \mbox{MSE ($\times10^3$) $\downarrow$} & \mbox{Pos.\ (mm) $\downarrow$} & \mbox{Rot.\ (deg) $\downarrow$} & \mbox{Grip $\downarrow$} & \mbox{Approach $\downarrow$} & \mbox{Grasp $\downarrow$} & \mbox{Align $\downarrow$} & \mbox{Release $\downarrow$}\\
    \midrule
    \rowcolor[RGB]{225,239,250}
    G/G & 0.947851 & 0.1090 & 0.03998 & \textbf{0.00841} & \textbf{0.016246} & \textbf{0.300583} & 2.898689 & 0.575885\\
    A/G & \textbf{0.861627} & \textbf{0.1034} & \textbf{0.03863} & \textbf{0.00841} & 0.060888 & 0.324025 & \textbf{2.485711} & 0.575885\\
    G/A & 1.037262 & 0.1090 & 0.03998 & 0.04193 & 0.139274 & \textbf{0.300583} & 3.171612 & \textbf{0.537579}\\
    A/A & 0.951039 & \textbf{0.1034} & \textbf{0.03863} & 0.04193 & 0.183916 & 0.324025 & 2.758634 & \textbf{0.537579}\\
    \bottomrule
  \end{tabular}
  \par\smallskip
  \begin{tabular}{@{}p{0.12\textwidth}>{\raggedleft\arraybackslash}p{0.14\textwidth}>{\raggedleft\arraybackslash}p{0.12\textwidth}>{\raggedleft\arraybackslash}p{0.10\textwidth}*{4}{>{\raggedleft\arraybackslash}p{0.13\textwidth}}@{}}
    \toprule
    \multicolumn{8}{l}{\textbf{(C) Frozen-input dependence: overall and phase-resolved errors}}\\
    & \multicolumn{3}{c}{Overall errors} & \multicolumn{4}{c}{Phase MSE ($\times10^3$)}\\
    \cmidrule(lr){2-4}\cmidrule(lr){5-8}
    Input & \mbox{MSE ($\times10^3$) $\downarrow$} & \mbox{Pos.\ (mm) $\downarrow$} & \mbox{Grip $\downarrow$} & \mbox{Approach $\downarrow$} & \mbox{Grasp $\downarrow$} & \mbox{Align $\downarrow$} & \mbox{Release $\downarrow$}\\
    \midrule
    \rowcolor[RGB]{225,239,250}
    Full & \textbf{0.947851} & \textbf{0.1090} & 0.00841 & 0.016246 & 0.300583 & \textbf{2.898689} & 0.575885\\
    No rich & 7.295912 & 0.1804 & 0.07862 & 0.019789 & \textbf{0.269088} & 4.363805 & 24.530964\\
    No events & 0.976991 & 0.1099 & \textbf{0.00680} & \textbf{0.009989} & 0.355350 & 2.990882 & \textbf{0.551741}\\
    Neither & 7.487956 & 0.1953 & 0.08005 & 0.012198 & 0.472941 & 5.537737 & 23.928948\\
    \bottomrule
  \end{tabular}
  \caption{\textbf{Ablation study of tactile refresh, intervention, and input features.}
  All conditions use the same frozen model and recorded histories.
  (A) varies refresh timing and regional reuse;
  (B) varies pose/gripper intervention; (C) masks tactile inputs.
  Light blue marks Event/Local, G/G, and full input as references.
  Bold marks each column's minimum within a panel, including ties.
  Requests and Calls count scheduled refreshes and encoder invocations, respectively.
  Encoder and Proc.\ give cumulative encoding and tactile-processing time;
  Enc.\ is mean encoding time per call.
  P/G denotes pose/gripper: G is gated and A always on.
  No rich and No events mask encoded tactile features and event summaries,
  respectively; Neither masks both.
  Pos./Rot. are translation/rotation command-component RMSE;
  Grip is gripper MAE in native $[-1,1]$ command units.
  Approach, Grasp, Align, and Release denote approach/normal,
  grasp correction, alignment/placement, and release/withdrawal.
  Overall MSE weights the four phases equally.}
  \label{tab:ablations}
  \label{tab:phase_ablations}
\end{table*}

%% file: sec/fig_core_analysis.tex
\begin{figure*}[t]
  \centering
  \includegraphics[width=\textwidth]{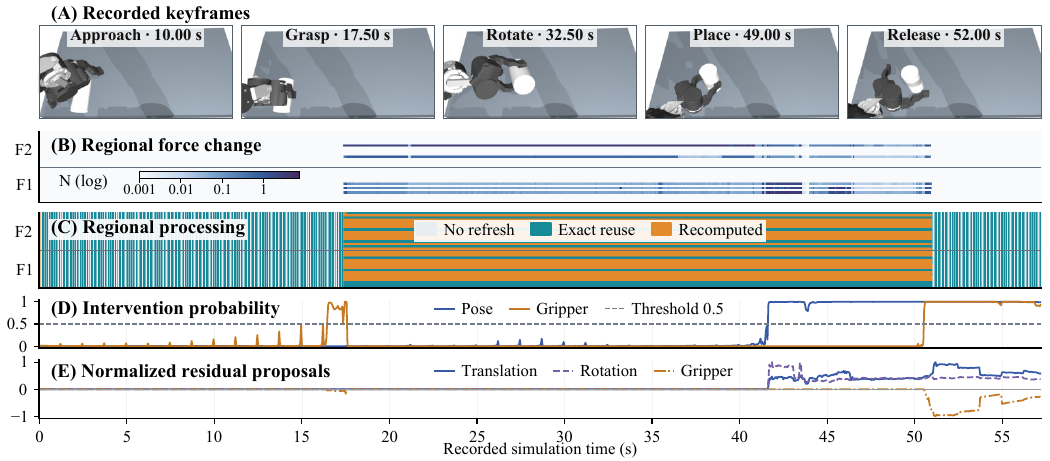}
  \caption{\textbf{Regional refresh and intervention are separate decisions.}
  (A) Recorded keyframes; (B) regional 1\,ms force-change peaks
  per 50\,ms bin (N, log); (C) processing states;
  (D) pose/gripper gate probabilities; (E) unexecuted gated proposals.
  F1/F2 cover regions 1--16/17--32.
  Panel C shows dependency-aware states, with recomputation
  overriding reuse; states are not call counts.
  Panel E uses separate absolute maxima: 2.17\,mm translation,
  0.771\,deg rotation, and 1.84 native gripper units (signed).
  Offline replay uses the frozen model in Table~\ref{tab:ablations},
  assumes immediate completion, and excludes the release-hold rule.}
  \label{fig:core_analysis}
\end{figure*}

%% file: sec/fig_physical_validation.tex
\begin{figure*}[t]
  \centering
  \includegraphics[width=\textwidth]{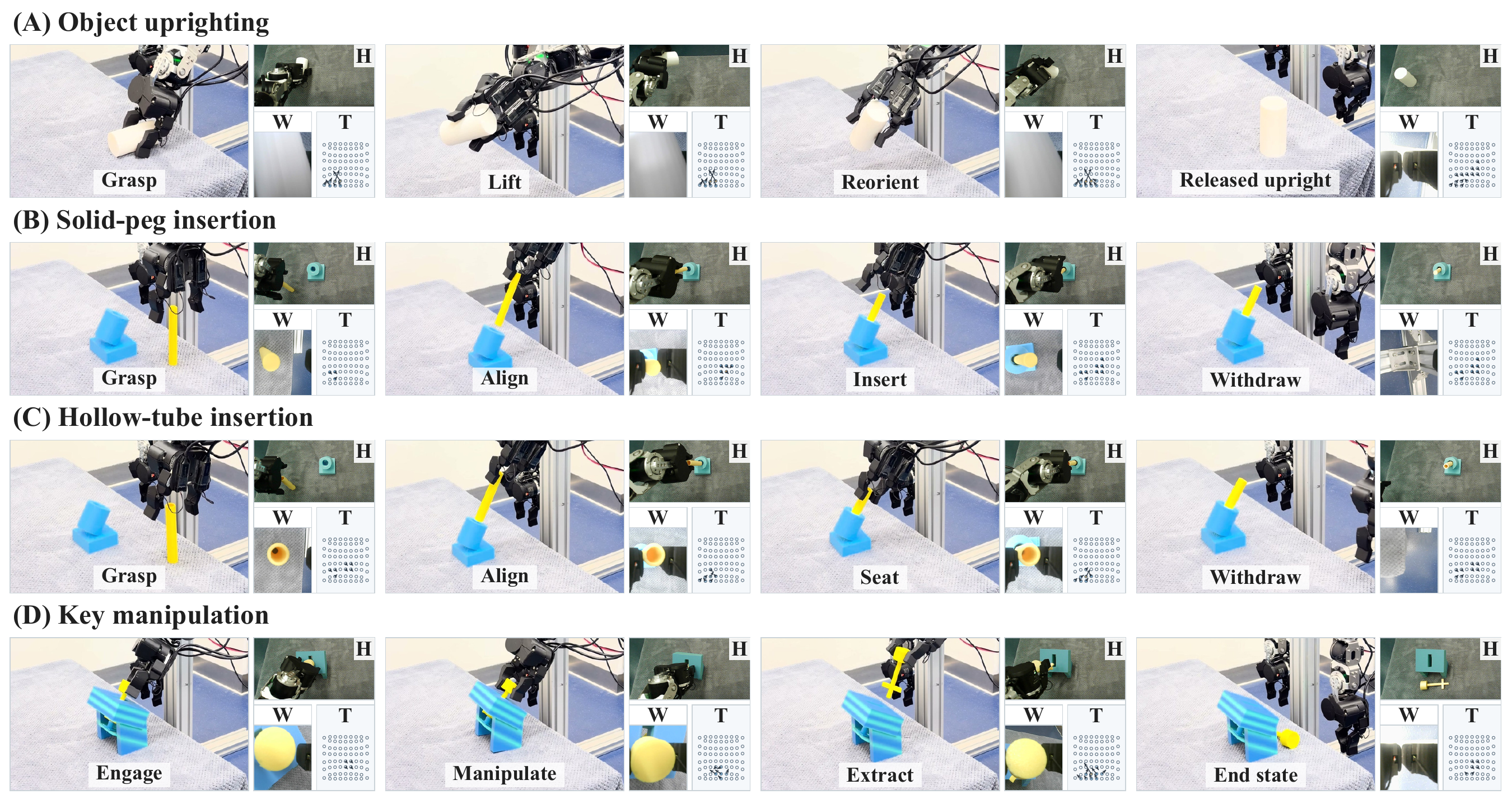}
  \caption{\textbf{Four contact-rich physical tasks on OpenArm v2.}
  Representative stages show (A) cylinder uprighting,
  (B) solid-peg insertion, (C) hollow-tube insertion, and
  (D) key manipulation. Each stage pairs an external view with
  head RGB (H), left-wrist RGB (W), and left-pad tactile measurements (T).
  Color and arrows show raw normal and shear sensor counts in
  sensor-local coordinates. Views use approximate event alignment
  ($\pm2$\,s; $\pm3$\,s for C); tactile values are uncalibrated counts.}
  \label{fig:physical}
\end{figure*}

%% file: sec/5_conclusion.tex
\section{Conclusion}
\label{sec:conclusion}

We presented TacPilot, which uses the spatial locality and
temporal intermittency of tactile changes to separately
control feature updates and corrections to a frozen visual policy.
Event-triggered regional updates reuse unchanged features,
while independent pose and gripper gates control bounded
Cartesian corrections. Cylinder-uprighting development trials
yielded 26/30 successes across two sizes. Offline analysis
showed 56.4\% fewer encoded regions with essentially unchanged
predictions; qualitative real-world experiments covered four
contact-rich tasks. System differences limit attribution
of baseline results to gating alone.
Future work will explore learning where and when to process
touch, and when to intervene, according to task needs.